\documentclass[a4paper,conference]{IEEEtran}
\IEEEoverridecommandlockouts
\usepackage{cite}
\usepackage{amsmath,amssymb,amsfonts,pgfplots,subfigure,multirow,booktabs}

\usepackage{array,enumitem}
\newcolumntype{P}[1]{>{\centering\arraybackslash}m{#1}}
\newcolumntype{L}[1]{>{\arraybackslash}m{#1}}

\usepackage{algorithmic}
\usepackage{graphicx}
\usepackage{textcomp}
\usepackage{xcolor}
\usepackage{url}
\usepackage{hyperref}
\def\BibTeX{{\rm B\kern-.05em{\sc i\kern-.025em b}\kern-.08em
    T\kern-.1667em\lower.7ex\hbox{E}\kern-.125emX}}
\begin{document}

\title{Exploiting Local Indexing and Deep Feature Confidence Scores for Fast Image-to-Video Search
}

\author{\IEEEauthorblockN{Savas Ozkan, Gozde Bozdagi Akar}
\IEEEauthorblockA{\textit{Department of Electrical/Electronics Engineering, Middle East Technical University} \\ \textit{06800, Ankara, Turkey}}
}

\maketitle

\begin{abstract}
The cost-effective visual representation and fast query-by-example search are two challenging goals that should be maintained for web-scale visual retrieval tasks on moderate hardware. This paper introduces a fast and robust method that ensures both of these goals by obtaining state-of-the-art performance for an image-to-video search scenario. Hence, we present critical enhancements to well-known indexing and visual representation techniques by promoting faster, better and moderate retrieval performance. We also boost the superiority of our method for some visual challenges by exploiting individual decisions of local and global descriptors at query time. For instance, local content descriptors represent copied/duplicated scenes with large geometric deformations such as scale, orientation and affine transformation. In contrast, the use of global content descriptors is more practical for near-duplicate and semantic searches. Experiments are conducted on a large-scale Stanford I2V dataset. The experimental results show that our method is useful in terms of complexity and query processing time for large-scale visual retrieval scenarios, even if local and global representations are used together. The proposed method is superior and achieves state-of-the-art performance based on the mean average precision (MAP) score of this dataset. Lastly, we report additional MAP scores after updating the ground annotations unveiled by retrieval results of the proposed method, and it shows that the actual performance.

\end{abstract}

\begin{IEEEkeywords}
Feature Indexing, Deep Features, Visual Content Retrieval, Image-to-Video Search
\end{IEEEkeywords}

\section{Introduction}
In the last decades,  we have witnessed an unprecedented proliferation of web-based multimedia data. The high-data population aggravates to retrieve a query from an extensive multimedia collection with a moderate hardware configuration.

Existing works\cite{mikolajczyk2005,sivic2003,philbin2007,yi2016,krizhevsky2012,simonyan2014very,bebanko2015,arandjelovic2016,gordo2016} primarily focus on two goals: high performance and fast query. Visual representations computed in an offline step are not marked as one of the goals since it is done before the query is made. However, a fundamental question that needs to be addressed is that \textit{'Is it enough to obtain the highest or fastest accuracy to deploy a complete retrieval system for users?'}. A plausible answer should be that, especially for larger multimedia databases (i.e., approaching real-world scenarios), solutions must consider hardware limitations before the querying to mitigate offline step complexity.

In this work, we present a visual multimedia retrieval method that aims to obtain high retrieval accuracy while keeping content presentation/indexing compact. Our main contributions are as follows:

\begin{itemize}[leftmargin=0.4cm,rightmargin=0cm]

\item In this work, we use local and global visual features together. This step plays a crucial role by solving the weaknesses of each feature set with the superiority of other methods. More specifically, local features tend to perform better with severe scale, rotation, and translation changes~\cite{lowe2004,mikolajczyk2005}. Similarly, global features can provide better results for semantic tasks because part-based visual representations are utilized~\cite{perronnin2010,krizhevsky2012,simonyan2014very}. For this purpose, we combine the confidence scores of local and global features detected for the same scenes with a novel fusion technique. The direct fusion of confidence scores is not appropriate. Hence, our method aims to normalize the confidence scores of both local and global features first. It then merges them into a final ranking list by adaptively selecting a settling point for each query.

\begin{figure*}[t]

\centering
\includegraphics[scale=0.28]{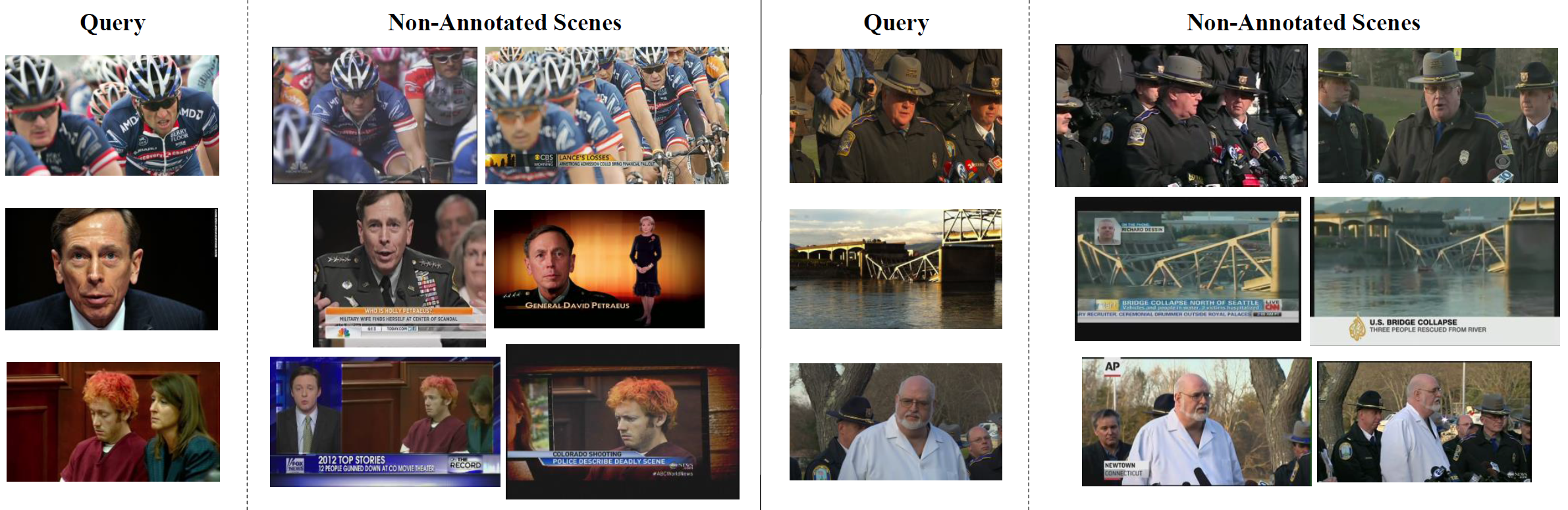}

  \caption{Non-annotated scene samples are unveiled by our retrieval results on Stanford I2V dataset. Our method can retrieve queries with severe viewpoint and conditional changes at the top of the ranked list.}
\label{fig:sample}
\end{figure*}

\item  In particular, the utilization of local and global descriptors for retrieval scenarios may conflict with low-cost computation limitations. Hence, we introduce improvements to well-known indexing and global pooling techniques; namely, Product Quantization (PQ)~\cite{jegou2011} and Compressed Fisher Vector (CFV) ~\cite{duan2014}, to balance the workload and to promote modest visual description. In short, we propose a non-parametric weighting function to compute probabilistic similarity scores between local features for query and reference data in an asymmetric PQ space. Furthermore, we replace hand-crafted features~\cite{lowe2004} and sparse keypoints ~\cite{lowe2004,mikolajczyk2005} with densely sampled mid-level convolutional features. Notice that this step still has low computation workload, since deep features are densely estimated. In addition, semantic content can be represented precisely with deep features~\cite{krizhevsky2012,bebanko2015}, which differs from the local content (relevant to the goal of our work). Last but not least, we apply an approximate binary nearest neighboring search (NN) to make querying operation up to 6x faster for CFV with a minor decrease for accuracy.

\item To this end, the proposed method enables fast query and low computational workloads for large-scale datasets while it outperforms the baselines by a large margin. Moreover, ground truth annotations for Stanford I2V dataset are updated that allow us to have a more reliable performance evaluation for future works.

\end{itemize}
\section{Related Work}
Here, we review fundamentals and most recent studies related to visual content search tasks. 

\noindent
\textbf{Local and Global Descriptors.} With the advent of sparse local features~\cite{lowe2004,mikolajczyk2005}, this idea instantly becomes popular for the visual representation domain. However, the usage of these features directly is not possible due to their dimensionality and large body. Capturing perfect descriptor relations by partitioning feature space into multiple clusters is a pioneering technique to ease this limitation~\cite{sivic2003}. However, to achieve discriminative hash codes, cluster size should be quite large even if an approximate search is utilized~\cite{philbin2007}.

Product quantization ~\cite{jegou2011,ge2014} reduces the cost of feature space partitioning significantly and mapping, since complexity is decreased by splitting a feature vector into multiple sub-vectors. Recently, the studies begin to focus on the adaptation of end-to-end learning techniques. Yi et al~\cite{yi2016} propose a deep network pipeline for keypoint detection, orientation estimation and visual description in a unified manner. However, the main problem is that relatively high hardware configurations are needed to complete computations in an acceptable time. Recently, the studies concentrate on both computation and performance constraints~\cite{martinez2018}.

On the other hand, orderless feature pooling (i.e., global descriptors) is an essential technique for retrieval tasks to produce a description from an independent set of features~\cite{perronnin2010}. Similarly, large dimensionality aggravates the direct use of these descriptors. Although PCA-like methods can be used to transform descriptors to several principal axes, binary codes that are calculated with a simple threshold~\cite{duan2014} provide several advantages as explained in~\cite{cevikalp2017}. Recently, neural networks are also used to compress feature vectors for effective search~\cite{morozov2019}.

Similarly, Babenko et al.~\cite{bebanko2015} compute deep features from different fragments of images and aggregate them with the VLAD descriptor. NetVlad~\cite{arandjelovic2016} is a trainable generalization of VLAD to recognize the geo-location of images. Gordo et al.~\cite{gordo2016} improve performance on the visual landmark retrieval task by finetuning a deep model with a Siamese triplet loss.  Lastly, ~\cite{araujo2018} shows that local and global descriptors should be jointly used to eliminate possible outliers for accurate retrieval. However, note that these methods need labeled data to train/finetune high-level features for deep models, that is impractical for all datasets.

\noindent
\textbf{Low Offline Workload.} As stated, a limited number of studies primarily consider low computational workload objectives in the literature~\cite{duan2014,araujo2014}. These solutions usually rely on representing a scene with a global visual descriptor. Also, the content around sparsely sampled points is used to achieve a reasonable computational workload. However, these descriptors tend to capture the content heavily from the background and repetitive parts~\cite{sivic20132}. Moreover, the discriminative power of representations (i.e., for binary codes) can decrease exponentially when the database size is increased~\cite{araujo2017}. Therefore, the single use of global descriptors remains weak for large-scale data, and additional representations should be exploited without sacrificing computational workload too much. 

\section{Proposed Method}

Our main goal is to enhance the performance of visual retrieval tasks by exploiting the confidence scores of both local and global descriptors for the same scenes. Thus, the computational load should be low so that a large amount of data can be processed within a reasonable time. 

Since our model is formulated as an image-based framework, keyframes are sampled at the start from a sequence of video frames  $ V = \{{v_1, v_2, ..., v_n} \} $ with a simple heuristic rule. This rule enforces a uniform sampling strategy -1fps- and a constraint that each frame should not contain any sub-region with large motion variations. Otherwise, these frames are discarded. 

\subsection{Local Visual Content Representation}

The local visual content of an input frame is computed from sparsely sampled key points. Then, these samples are converted into compact hash codes. In short, a two-stage quantization-based approach is utilized. Moreover, geometric consistency between local features is added with fast geometric filtering at the end.

\noindent
\textbf{Local Sparse Features.} For local representation, we use Root SIFT~\cite{aran2012} and Hessian Laplacian~\cite{mikolajczyk2005}. Since they are robust to scale and orientation changes, we expect to successfully retrieve any query that presents strong geometric deformations (duplicate/copy) from the reference set.

In addition, to find geometric adjacency of local matches at the voting step, we store coordinate $ (x, y)  $, orientation  $\theta$ and scale $s$ coefficients of each local point in quantized forms.

\noindent
\textbf{Feature Indexing.} As mentioned, the direct use of local features is impractical and should be converted into small representations. The idea of Bag-of-Word-like~\cite{sivic2003} methods is to quantize  (Eq.~\ref{eqn:bc}) each feature vector  $ f_h\in \mathbb{R}^{128}$  (note that dimension of SIFT is 128 and k-means is used) to the closest center $c_i$ from a pre-clustered feature space $ C_{bow} \in \mathbb{R}^{D_{bow} \times 128}$:
\begin{eqnarray}
\label{eqn:bc}
q_b(f_h) =  \min_{i} \parallel f_h - c_i \parallel_2, c_i \in C_{bow},
\end{eqnarray}
However, the number of cluster centers should be adequately large (e.g. $ D_{bow} > 100K $) in order to achieve discriminative space partitions~\cite{philbin2007}. This requirement leads to an increase in offline calculations and makes it difficult to represent visual content with local features.

In our work, a rational computation effort is achieved by encoding residual vector ($ r=f_h-c_i$) with an additional quantizer $ q_b(.) $. Hence, relatively smaller cluster sizes (i.e., $ D_{bow} \approx 10K $) can be selected. PQ space~\cite{jegou2011} $ C_{pq} = \{C^1_{pq}, C^2_{pq}, ..., C^m_{pq}\},  C^k_{pq} \in \mathbb{R}^{D_{pq} \times 128/m }$  is selected to maximize information bit per component by splitting residual vector into $m$ non-overlapping sub-vectors $ r = \{r_1, r_2,$ $..., r_m\}$ as follows:
\begin{eqnarray}
\label{eqn:pqc}
q^{k}_{pq}(r) =  \min_{i} \parallel r_k - c_i \parallel_2, c_i \in C^k_{pq}, \forall k.
\end{eqnarray}

At the end, each feature vector is converted into two interrelated hash codes where $h_b = q_b$ and $h_{pq} = \{ q^1_{pq} , q^2_{pq}, ..., q^m_{pq} \} $, and they are stored in an inverted file structure based on their $h_b$ values. Throughout this work, $m$ and $D_{pq}$ are empirically set to 8 and 256 as in~\cite{jegou2011}. In a nutshell, computation cost is exponentially reduced with low cost two-stage indexing scheme and local representation becomes applicable for large-scale databases.

\noindent
\textbf{Local Voting Scheme.} We use a two-fold approach for the voting scheme: First, we estimate the best locally matching candidates based on similarities between hash codes. Later, outliers are eliminated based on the dominant geometric model between the query and reference frames.

Formally, in order to say that query and reference local points are similar, coarse hash codes should be the same ($h^q_b = h^r_b$), while residual similarities ($h^r_{pq}$ and $h^q_{pq}$) must be within an error tolerance. Otherwise, the similarity score is set to zero and discarded from initial query candidates.

This work proposes a novel non-parametric score function for PQ Euclidean space to be used in the residual similarity calculation. Ultimately, this score function normalizes asymmetric Euclidean distance of two residual hash codes with a maximum asymmetric distance between all cluster centers. Later, the final residual similarity score is equal to the average of all subvector scores:
\begin{eqnarray}
\label{eqn:pqs}
w_{pq}(h^r_{pq}, h^q_{pq}) = \dfrac{1}{m} \sum\limits_{k=1}^{m}  \bigg(1-\dfrac{1}{d_k}\parallel  q^{r,k}_{pq} - q^{q,k}_{pq} \parallel_2 \bigg). 
\end{eqnarray}

\noindent
where $d_k$ indicates the maximum asymmetric Euclidean distance for $k^{th}$ subvector, i.e., $d_k = \max{\parallel c_i - c_l \parallel_2}$, $\forall i, l \in D_{pq}, c_i, c_l \in C^k_{pq}$. This non-parametric function enables us to assess similarities within [0,1] ($w_{pq}(.,.) \in [0, 1]$ ) rather than varying Euclidean distances. To this end, a threshold-based coefficient is used to select best matches.

After initial similarity scores are obtained by using Eq. (\ref{eqn:pqs}), hard-similarity scores (these scores can be equal to either 0 or $w_{pq}$) are determined by applying a coarse threshold $\tau_{pq}$. Practically, this step improves our method by selecting the best matches that yield high confidence scores, and it removes possible outliers immediately. Moreover, we prune 5\% of the codewords according to their term frequencies to reduce the drawback of stop words~\cite{sivic2003} and to speed up the querying. Note that hard-similarity scores ($w_{pq}(.,.)>\tau_{pq}$ and $h^q_b = h^r_b$) are also weighted by this frequency term.

Later, we filter out outliers (i.e., the ones that not obey to dominant geometric model between the query and reference frames) by enforcing a 4-dof geometric constraint (affine transform might yield better results, yet it increases query time) (\ref{eqn:geo}) on initial query candidates: 
\begin{eqnarray}
\label{eqn:geo}
&\left ( \begin{matrix}
x^q\\ 
y^q\\ 
1
\end{matrix} \right )  = \begin{bmatrix} 
   \widetilde{s}\cos{\widetilde{\theta}} &  -\widetilde{s}\sin{\widetilde{\theta}} & t_x \\
   \widetilde{s}\sin{\widetilde{\theta}} &  \widetilde{s}\cos{\widetilde{\theta}} & t_y \\ 
   0 & 0 & 1 
\end{bmatrix} \left ( \begin{matrix}
x^r\\ 
y^r\\ 
1
\end{matrix} \right ).
\end{eqnarray}

\noindent
where $\widetilde{s} = s^q - s^r$ and $ \widetilde{\theta} = \theta^{q} -  \theta^{r} $ are the differences of scale and orientation parameters for query and reference points. ($t_x$, $t_y$) is also spatial translation between query and reference local points, and their values are estimated from Eq. \ref{eqn:geo}. This geometric model simply calculates a histogram using common parameter distributions of scale ($\log(\widetilde{s})$), orientation ($\widetilde{\theta}$) and translation (($t_x+t_y)/\widetilde{s}$) between query candidates of each frame. Later, the highest scored bin (sum of all hard-similarity scores that fit the dominant parameter distribution) yields the dominant geometric model between local points. To this end, the dominant value is set as a final local confidence score for a frame.

\subsection{Global Visual Content Representation}

Pretrained deep convolutional features are densely sampled and reduced by PCA to lower dimensions. Then, these features are aggregated with Fisher Kernel~\cite{perronnin2010} and transformed into binary hash codes. These binary codes are compared with an approximate NN search setting in Hamming space to ease the querying step.

\noindent
\textbf{Dense Deep Convolution Features.} We use densely sampled pre-trained deep convolutional features $ f_d\in \mathbb{R}^{384} $ obtained at Alexnet-conv3 layer~\cite{krizhevsky2012} by discarding zero paddings in convolution layers. As proved~\cite{krizhevsky2012,bebanko2015,simonyan2014very}, deep features depict the semantic content of a scene more precisely than hand-crafted features. By replacing hand-crafted features, we contain this semantic model with local structures to realize a complete retrieval method. Furthermore, we select conv3 features to keep the computational complexity low.

Later, these densely sampled features are mapped to 64-dimensional space by PCA. There are two main reasons: First, ~\cite{bebanko2015} shows that PCA-compressed deep features are robust since degrading the sparsity of features on a different visual set can improve their generalization capacity for unseen examples. Second, it provides time advantages in computations with feature pooling and voting stages. 

\noindent
\textbf{Feature Pooling and Fast Voting Scheme.} Deep features are aggregated with first-order Fisher Kernel~\cite{perronnin2010} to estimate one compact representation $ v\in \mathbb{R}^{64 \times D_{fk}}  $ for each frame ($D_{fk}$ is the number of Gaussian mixture components). Since dimensionality does not allow us to search and store them in large-scale databases, they are converted into binary codes $b$ by applying a zero-bias threshold rather than quantization-based approaches. The main reason to select a threshold-based approach is that~\cite{cevikalp2017} shows that Euclidean-based quantization can be misleading for high dimensional representations. We also prove this assumption in our experiments.

In addition, even if converting a high-dimensional descriptor into a compact binary code speeds up the query time, there is still room to further eliminate the redundant calculations for our method. We replace the standard brute-force binary search with an approximate nearest neighboring (NN) scheme in this work. 

In the proposed method, initial matched candidates for global binary codes are obtained based on KNN results in Hamming space (i.e., using inverted index structure the same as in local descriptors). Later, full distances (\ref{eqn:knn}) are computed by only comparing these candidates to improve confidence scores of query and reference frames.
\begin{eqnarray}
\label{eqn:knn}
&&w_{b}(b^r, b^q) = \Bigg\{ 
\begin{matrix}
g_h(b^r, b^q), & \text{\parbox{4cm}{if $b^r$ is in KNN of $b^q$}} \\ 
0, & \text{\parbox{4cm}{otherwise}} \\ 
\end{matrix}
\end{eqnarray}

\noindent
where $g_h(.,.)$ is the score function, which returns normalized Hamming distance for two binary codes (normalized by the total number of bits, and similarly probabilistic scores are generated). Moreover, binary space is partitioned into 32 cluster centers throughout experiments.

\subsection{Late Fusion} 

Until now, we compute a set of visual descriptors and construct two individual databases for local and global representations by depicting the same visual content. The most similar video scenes are retrieved separately by using these databases. Hence, we obtain two ranked lists, as illustrated in Fig.~\ref{fig:rank} for each query. From these lists, our objective is to obtain one final ranked list by fusing these confidence scores.

However, the fusion of these decisions is not straight-forward. Even if confidence scores are in a similar range $[0, 1]$, there is no common score characteristic that can be directly exploited for all queries. Hence, confidence scores should be normalized separately for each query before mapping these values to a final ranked list.

From these plots in Fig.~\ref{fig:rank}, saturation of confidence scores for ranked lists shares similar characteristic (i.e., $0.75-0.65\simeq0.1$ and $0.15-0.02\simeq0.13$). Hence, this information can be exploited to fuse the scores. Therefore, a settling point must be initially determined from each list in order to normalize the scores. It can be done in two ways: 1) the last element of a list can be selected as a settling point, and scores in each list can be normalized by normalizing according to this value. 2) an adaptive point is determined from each list by using inner relations of confidence scores. Indeed, the second assumption yields better results, since it does not depend on the number of elements in ranked lists. Similarly, as in the assumption of query expansion~\cite{chum2011}, scores reflect an error characteristic after some points, and no content correlation is expected between the query and reference frames. Hence, our technique is inspired by the assumption of query expansion.

For this purpose, we iteratively calculate first-order score derivatives between all two consecutive confidence scores (i.e., subtracting one point from another). We then obtain an adaptively selectable point where gradient converges to a minimal number $\epsilon$ (e.g., $\epsilon$=0.01) after a period (after 10 elements). This point is accepted as a settling point, and all scores are normalized by subtracting this value.

Later, normalized local and global scores are merged by regarding their highest scores. Hence, the final ranked list is able to preserve both local structure and semantic similarities for each query in a unified form.

\section{Experiments}

Our experiments are conducted on Stanford I2V~\cite{araujo2014}. This dataset is particularly suitable for our method since it contains a large volume of videos collected from diverse news video archives to illustrate its actual capacity. 

\begin{figure}
\subfigure{
\begin{tikzpicture} 
\begin{axis}[
    axis lines = left,
    ylabel style={yshift=-0.2cm},
    xlabel style={yshift=0.1cm},
    width=9cm, height=3.2cm,
    ymin={0.65}, ymax={0.78},
    ytick={0.65, 0.70, 0.75},
    legend pos=north west,
]

\addplot[
   color=blue,
   mark=otimes*,
    ]
    coordinates {
    (1,0.775)(2,0.725)(3,0.719)(4,0.714)(5,0.710)(6,0.708)(7,0.704)(8,0.691)(9,0.687)(10,0.663)(11,0.662)(12,0.662)(13,0.661)(14,0.660)(15,0.659)(16,0.659)(17,0.659)(18,0.659)(18,0.658)(19,0.658)(20,0.657)(21,0.657)(22,0.657)(23,0.657)(24,0.656)(25,0.656)(26,0.656)(27,0.656)(28,0.656)(29,0.656)(30,0.656)(31,0.655)(32,0.655)(33,0.655)(34,0.654)(35,0.654)(36,0.654)(37,0.654)(38,0.654)(39,0.654)(40,0.654)(41,0.654)(42,0.653)(43,0.653)(44,0.653)(45,0.653)(46,0.653)(47,0.653)(48,0.653)(49,0.652)(50,0.652)
    };

\addplot[
   color=red,
   mark=square*,
    ]
    coordinates {
    (30,0.656)
    };

\end{axis}
\end{tikzpicture}
}

\subfigure{
\label{fig:b}
\begin{tikzpicture} 
\begin{axis}[
    axis lines = left,
    ylabel style={yshift=-0.2cm},
    xlabel style={yshift=0.1cm},
    width=9cm, height=3.2cm,
yticklabel style={
        /pgf/number format/fixed,
        /pgf/number format/precision=5
},
    ymin={0.02}, ymax={0.16},
    ytick={0.02, 0.075, 0.15},
    legend pos=north west,
]

\addplot[
   color=green,
   mark=otimes*,
    ]
    coordinates {
	(1,0.1559)(2,0.054)(3,0.049)(4,0.043)(5,0.042)(6,0.042)(7,0.042)(8,0.042)(9,0.042)(10,0.042)(11,0.042)(12,0.042)(13,0.042)(14,0.034)(15,0.031)(16,0.031)(17,0.031)(18,0.031)(19,0.031)(20,0.031)(21,0.031)(22,0.031)(23,0.031)(24,0.031)(25,0.031)(26,0.028)(27,0.028)(28,0.028)(29,0.027)(30,0.027)(31,0.027)(32,0.027)(33,0.027)(34,0.027)(35,0.026)(36,0.026)(37,0.026)(38,0.026)(39,0.026)(40,0.025)(41,0.024)(42,0.024)(43,0.024)(44,0.023)(45,0.023)(46,0.023)(47,0.023)(48,0.023)(49,0.022)(50,0.022)
    };

\addplot[
   color=red,
   mark=square*,
    ]
    coordinates {
	(24,0.031)
    };
 
\end{axis}
\end{tikzpicture}
}
\caption{\small Top confidence scores for two ranked lists obtained by global (blue) and local (green) descriptors. Red points indicate the adaptive settling points determined from each list.}

\label{fig:rank}
\end{figure}

\begin{figure}[b!]
\centering     
\subfigure{
\label{fig:list}
\begin{tikzpicture} 
\begin{axis}[
    axis lines = left,
    xlabel = \small $k$,
    ylabel = {\small $mAP (\%)$},
    ylabel style={yshift=-0.4cm},
    xlabel style={yshift=0.1cm},
    xticklabel style = {font=\small},
    yticklabel style = {font=\small},
    width=7cm, height=4cm,
    xmin=2.5, xmax=6.5,
    ymin=0.0, ymax={0.7},
    xtick={3, 4, 5, 6},
    ytick={0.1, 0.2, 0.3, 0.4, 0.5, 0.6},
    legend pos=north west,
    legend style={nodes={scale=0.9, transform shape}},
    ymajorgrids=true,
    grid style=dashed,
]

\addplot[
   color=red,
    mark=otimes*,
    ]
    coordinates {
    (3, 0.099)(4,0.191)(5,0.483)(6,0.496)
    };

\addplot[
   color=blue,
   mark=square*,
    ]
    coordinates {
    (3, 0.111)(4,0.209)(5,0.52)(6,0.542)
    };

\addplot[
   color=green,
   mark=triangle*,
    ]
    coordinates {
    (3, 0.121)(4,0.236)(5,0.541)(6,0.552)
    };

\legend{D$_{fk}$=64, D$_{fk}$=128, D$_{fk}$=256}
 
\end{axis}
\end{tikzpicture}}
\subfigure{
\label{fig:b}
\begin{tikzpicture} 
\begin{axis}[
    axis lines = left,
    xlabel = \small $k$,
    ylabel = {\small $mAP (\%)$},
    ylabel style={yshift=-0.4cm},
    xlabel style={yshift=0.1cm},
    xticklabel style = {font=\small},
    yticklabel style = {font=\small},
    width=7cm, height=4cm,
    xmin=2.5, xmax=6.5,
    ymin=0.0, ymax={0.7},
    xtick={3, 4, 5, 6},
    ytick={0.1, 0.2, 0.3, 0.4, 0.5, 0.6},
    legend pos=north west,
    legend style={nodes={scale=0.9, transform shape}},
    ymajorgrids=true,
    grid style=dashed,
]

\addplot[
   color=red,
   mark=otimes*,
    ]
    coordinates {
    (3, 0.090)(4,0.167)(5,0.394)(6,0.411)
    };

\addplot[
   color=blue,
   mark=square*,
    ]
    coordinates {
    (3, 0.091)(4,0.175)(5,0.452)(6,0.458)
    };

\addplot[
   color=green,
   mark=triangle*,
    ]
    coordinates {
   (3, 0.110)(4,0.206)(5,0.450)(6,0.463)
    };

 
\end{axis}
\end{tikzpicture}
}

\caption{Impact of k values for binary NN voting. Results for light (upper) and full (lower) sets are reported. }
\label{fig:global}
\end{figure}
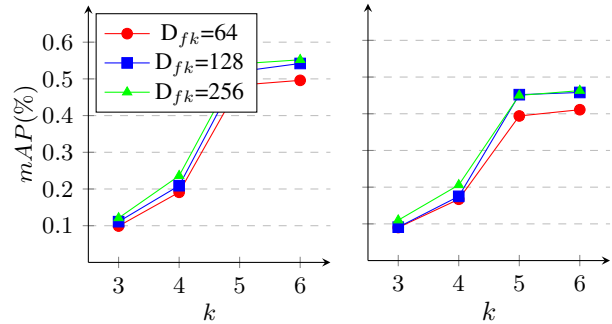

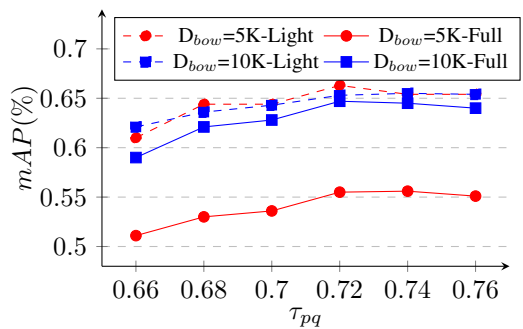
\begin{figure}[b]
\label{fig:pltae}
\begin{tikzpicture} 
\begin{axis}[
    axis lines = left,
    xlabel = $\tau_{pq}$,
    ylabel = {$mAP (\%)$},
    ylabel style={yshift=-0.2cm},
    xlabel style={yshift=0.1cm},
    width=9cm, height=5.0cm,
    xmin=0.65, xmax=0.77,
    ymin=0.48, ymax={0.74},
    xtick={0.66, 0.68, 0.7, 0.72, 0.74, 0.76},
    ytick={0.5, 0.55, 0.6, 0.65, 0.7},
    legend pos=north west,
    legend columns=2, 
    legend style={nodes={scale=0.8, transform shape}},
    ymajorgrids=true,
    grid style=dashed,
]

\addplot[
   color=red,
   dashed, mark=otimes*,
    ]
    coordinates {
    (0.66, 0.61)(0.68, 0.644)(0.70, 0.644)(0.72,0.663)(0.74,0.654)(0.76, 0.654)
    };
\addlegendentry{D$_{bow}$=5K-Light}

\addplot[
   color=red,
    mark=otimes*,
    ]
    coordinates {
    (0.66, 0.511)(0.68, 0.53)(0.70, 0.536)(0.72,0.555)(0.74,0.556)(0.76,0.551)
    };
\addlegendentry{D$_{bow}$=5K-Full}

\addplot[
   color=blue,
   dashed,mark=square*,
    ]
    coordinates {
    (0.66, 0.621)(0.68, 0.636)(0.70, 0.643)(0.72,0.653)(0.74,0.655)(0.76, 0.654)
    };
\addlegendentry{D$_{bow}$=10K-Light}

\addplot[
   color=blue,
   mark=square*,
    ]
    coordinates {
    (0.66, 0.59)(0.68, 0.621)(0.70, 0.628)(0.72,0.647)(0.74,0.645)(0.76,0.640)
    };
\addlegendentry{D$_{bow}$=10K-Full}

 
\end{axis}
\end{tikzpicture}
\caption{Impact of $\tau_{pq}$ for top 100 retrieved scenes. Results are reported for local descriptors.}
\label{fig:local}
\end{figure}

As stated in~\cite{araujo2014}, Stanford I2V dataset is split into two versions, such that a lighter version contains a subset of query images and reference videos of a full version. Full and light versions consist of 3801 and 1035 hours of videos, respectively (Please do not confuse SI2V-4M or SI2V-600K in~\cite{araujo2017}). Moreover, the amount of query images is decreased by factor 3, from 229 to 78. 

The provided script measures the performance in the evaluation step, and the mean Average Precision (MAP) scores are reported.
\begin{table}
\caption{Approximate time (sec) spent on the representation stage per frame on a single CPU core.}

\begin{tabular}{P{1.8cm} P{1.8cm} P{1.8cm} P{1.8cm}}
  \hline
    \multicolumn{4}{l}{\small Local Descriptor} \\
    \hline
     \small Keypoint & \small Descriptor  & \small Indexing & \small Total \\
    \small 0.223  & \small 0.410  & \small 1.331 & \small 1.996 \\
  \hline \hline
     \multicolumn{4}{l}{\small Global Descriptor} \\
  \hline
    \small Descriptor  & \small PCA & \small Fisher Kernel & \small Total \\ 
    \small 1.163  & \small 0.005  & \small 0.193 & \small 1.187 \\
  \hline

\end{tabular}
\label{tab:comp}
\end{table}

\noindent
\textbf{Computation Load.} The main objective is to accomplish a method that computation load and query processing time are moderate while obtaining state-of-the-art retrieval accuracy for large-scale data. Table~\ref{tab:comp} illustrates average time requirements per frame. Observe that it is close to real-time (assume that 1fps keyframe is processed). This feature allows us to analyze the visual content of this dataset within several days on cheap CPU servers. In similar, memory requirement is negligible since all descriptors are converted to compact hash codes.

\noindent
\textbf{Impact of the Parameters.} Our framework is composed of local and global ranking stages. Hence, first, we need to obtain the best parameters empirically for each stage. 

For local visual representations, $\tau_{pq}$ defines error tolerance for PQ signature matches. For small values, noisy versions of valid signatures can be estimated correctly. However, this reduces the discriminative power of signatures and introduces lots of outliers. Therefore, we set $D_{bow}$ as 5K and 10K to obtain trade-offs in terms of retrieval accuracy and query processing time for $\tau_{pq}$. Fig.~\ref{fig:local} illustrates mAP scores for top 100 ranked scenes based on various $\tau_{pq}$ values. From the results, settling $\tau_{pq}$ around 0.72 yields the best performance for all configurations. Another important observation is that 5K scores drop down drastically for the full version of this dataset. This result can be demonstrated by the fact that the discriminative power of this representation is not adequate for smaller values of $D_{bow}$. Moreover, a larger cluster size $D_{bow}$ ultimately provides an advantage in the querying stage by reducing operations due to an inverted index structure. 

\begin{table*}[t!]
\begin{center}
\caption{Late fusion results on SI2V dataset. Latency (sec) is measured for 1000h reference videos per query. Best results in each part are in bold.}
\begin{tabular}{P{3.0cm} P{1.2 cm} P{1.2 cm}P{1.2 cm} P{1.2 cm} P{2.0cm}}
  \hline  \toprule
  & \multicolumn{2}{c}{\small Light Dataset} & \multicolumn{2}{c}{\small Full Dataset} & \multirow{2}{1.5cm}{\centering\small Latency Per 1000h} \\ 

 {\small [D$_{bow}$-D$_{fk}$]}  & \small mAP & \small {mAP@1} & \small mAP & \small mAP@1 \\   
  \toprule
    \small EH~\cite{de2016large}   & \small -  & \small -  & \small 0.15 & \small  0.37 \ & \small - \\
  \small PHOG~\cite{de2016large}   & \small -  & \small -  & \small 0.22 & \small  0.45 \ & \small - \\
  \small SCFV~\cite{araujo2014}   & \small 0.46  & \small 0.73  & \small 0.43 & \small  0.64 \ & \small 12.75 sec \\
  \small BF-PI~\cite{araujo2017}  & \small  $\approx$0.68  & \small -  & \small $\approx$0.65 & \small -  \ & \small  $\approx$ 4.3 sec \\
  \small RMAC~\cite{garcia2018asymmetric}  & \small  -  & \small -  & \small $\approx$0.66 & \small -  \ & \small  - \\ 
    \hline  
  \hline

  \small ours[5K - 64]    & \small 0.667  & \small 0.769  & \small 0.582  & \small 0.716 & \small 17.11 sec \\
  \small ours[5K - 128]  & \small 0.695  & \small \textbf{0.794}  & \small 0.601  & \small 0.755 & \small 18.237 sec \\
  \small ours[5K - 256]  & \small \textbf{0.707}  & \small 0.782  & \small 0.622  & \small 0.755 & \small 19.253 sec \\

  \small ours[10K - 64]    & \small 0.668  & \small 0.769 & \small 0.644  & \small 0.764 & \small 8.675 sec \\
  \small ours[10K - 128]  & \small 0.679  & \small 0.782 & \small 0.663  & \small \textbf{0.786} & \small 9.802 sec \\
  \small ours[10K - 256]  & \small 0.700  & \small 0.782 & \small \textbf{0.670}  & \small 0.777 & \small 10.809 sec \\

  \hline
\end{tabular}
\label{tab:t1}

\end{center}
\end{table*}

For global visual representations, $D_{fk}$ and $k$ are two parameters that users need to define. We select 64, 128 and 256 for the number of Gaussian mixture components (D$_{fk}$). Since $k$ value determines redundancy in approximate binary code search, accuracy and query processing time are influenced inversely from this value. The accuracy saturates at $k=5$ for all configurations, as shown in Fig.~\ref{fig:global}. This configuration speeds up search time approximately 6x faster. As expected, increasing the number of mixture components ($D_{fk}$) restores accuracy for both versions. However, the total number of comparisons, as well as storage requirements, becomes more extensive.

\noindent
\textbf{Impact of Late Fusion.} We calculate mAP scores after fusing confidence scores of local and global descriptors in various combinations. Table~\ref{tab:t1} shows that late fusion boosts retrieval accuracy around 5\% compared to their individual baselines estimated by local and global representations (i.e., Fig.~\ref{fig:global} and Fig.~\ref{fig:local}). Also, the combination of 5K-256 obtains the best mAP accuracy for the light set. However, scores are decreased for the full version due to the failure of PQ. On the other hand, 10K-256 combination yields compatible scores for both full and light versions.

In the querying stage, the combination of 10K-64 yields the fastest response time. Since global representations are also stored in an inverted index structure, the use of high-dimensional representation increases the sparsity of codewords. The number of computations is reduced compared to 5K representation.

\noindent
\textbf{Baselines.} We compare our performance with the reported baseline results in the literature (Table~\ref{tab:t1}). Initial database performance~\cite{araujo2014} on light and full versions are approximately 46\% and 43\% for mAP@100. Later, even if worse performance is achieved, authors present simple yet somehow effective global representations~\cite{de2016large}. More recently,~\cite{araujo2017} achieves an additional 21\% mAP improvement compared to baseline scores by using a shot-based feature aggregation technique. RMAC based deep feature pooling is also adopted~\cite{garcia2018asymmetric}. Lastly, the latency of~\cite{araujo2017} might be better than our method due to promoting frame-based assumption to shot-based assumption. However, remark that shot-based features introduce additional workloads for offline computations. Notice that our method obtains state-of-the-art performance on SI2V dataset.

\noindent
\textbf{Updating the Ground Truth Annotations.}  We update ground truth annotations for both full and light versions of SI2V dataset. These annotations are unveiled with our retrieval results (\url{https://github.com/savasozkan/i2v}). 

Annotation pipeline of SI2V dataset~\cite{araujo2014} relies on an automated annotation process, as explained by the authors. Precisely, reference videos candidates are initially pruned with a time constraint (based on time tag of queries), and a feature-based matching technique is utilized before any human visual intervention. As a result, some of the scenes might be discarded unintentionally from annotation lists. Therefore, we manually examine our top retrieval results (up to 20 scenes per query) and find out that some of the retrieved results are non-annotated in the ground truth, even if they have strong semantic analogies and visual copies with query images. We illustrate some of the scene samples in Fig.~\ref{fig:sample}. 

Table~\ref{tab:t2} shows mAP scores recalculated with our updated ground truth annotations. The results introduce additional 5\% improvements compared to Table~\ref{tab:t1} for light version of dataset. This result is profoundly critical since the actual performance of our method is even beyond the reported performance in the literature. Moreover, although there is a noticeable performance increase for the light set, this increase is not as much for the full version. The reason is that the selection capacity of global binary representation saturates for the larger set, as explained. This notion validates the importance of the joint use of local and global representations for retrieval tasks. To make fair comparisons, we also implement and test the baseline method proposed in~\cite{araujo2014,de2016large} on updated ground truth annotations. From the results, it provides only 2\% and 1\% additional improvement for~\cite{araujo2014} while 4\% improvement is obtained for~\cite{de2016large}.

\begin{table}[t!]
\label{tab:load}
\begin{center}

\caption{Late fusion mAP after updating the ground truth annotations. Best results in each part are in bold.}

\begin{tabular}{P{1.9cm} P{1.2 cm} P{1.2 cm}P{1.2 cm} P{1.2cm}}
  \hline  \toprule
  & \multicolumn{2}{c}{\small Light Dataset} & \multicolumn{2}{c}{\small Full Dataset} \\ 

 {\small [D$_{bow}$-D$_{fk}$]}  & \small mAP & \small {mAP@1} & \small mAP & \small mAP@1  \\   
  \toprule
  \small [5K - 64]    & \small 0.697  & \small 0.794  & \small 0.577 & \small 0.720 \\
  \small [5K - 128]  & \small 0.735  & \small 0.833  & \small 0.607 & \small 0.755  \\
  \small [5K - 256]  & \small \textbf{0.755}  & \small \textbf{0.846}  & \small 0.624  & \small 0.764  \\

  \small [10K - 64]    & \small 0.708  & \small 0.807 & \small 0.648  & \small 0.768  \\
  \small [10K - 128]  & \small 0.729  & \small 0.820 & \small 0.667  & \small 0.786  \\
  \small [10K - 256]  & \small \textbf{0.755}  & \small 0.833 & \small \textbf{0.681}  & \small \textbf{0.790}  \\
  \hline  
  \hline
  \small EH~\cite{de2016large}   & \small -  & \small -  & \small 0.19 & \small  0.42 \\
  \small SCFV~\cite{araujo2014}   & \small 0.48  & \small 0.76  & \small 0.44 & \small  0.68 \\
  \hline

\end{tabular}
\label{tab:t2}

\end{center}
\end{table}

\section{Conclusion}

In this work, we introduce a visual search method for large-scale visual retrieval task. It exploits local and global descriptors together to represent visual data. The primary objective of the proposed method is to obtain moderate computation load and query time for large-scale datasets. Furthermore, performance is improved compared to baselines. We present critical contributions to the techniques for visual representation and feature hashing throughout the paper. In addition, we propose a novel technique to fuse local feature-based scores and deep global scores as a late fusion step. To show the superiority of our method, experiments are conducted on Stanford I2V dataset. As explained, it achieves the state-of-the-art mAP performance in the literature. Moreover, we update ground truth annotations for Stanford I2V based on the retrieval results of the proposed method. The final results show that the actual performance of our method is much better after updated ground truth annotations are used.

\section*{Acknowledgment}

The authors gratefully acknowledge the support of NVIDIA Corporation with the donation of GPUs used for this research.

\bibliographystyle{IEEEbib}
\bibliography{icme2020template}

\begin{thebibliography}{10}

\bibitem{mikolajczyk2005}
K.~Mikolajczyk and C.~Schmid.,
\newblock ``A performance evaluation of local descriptors.,''
\newblock {\em IEEE transactions on pattern analysis and machine intelligence},
  pp. 1615--1630, 2005.

\bibitem{sivic2003}
J.~Sivic and A.~Zisserman,
\newblock ``Video google: A text retrieval approach to object matching in
  videos,''
\newblock {\em Proceedings of the IEEE international conference on computer
  vision}, pp. 1470--1477, 2003.

\bibitem{philbin2007}
J.~Philbin, O.~Chum, M.~Isard, J.~Sivic, and A.~Zisserman,
\newblock ``Object retrieval with large vocabularies and fast spatial
  matching,''
\newblock {\em Proceedings of the IEEE Conference on Computer Vision and
  Pattern Recognition}, pp. 1--8, 2007.

\bibitem{yi2016}
K.M. Yi, E.~Trulls, V.~Lepetit, and P.~Fua,
\newblock ``Lift: Learned invariant feature transform,''
\newblock {\em European conference on computer vision}, pp. 467--483, 2016.

\bibitem{krizhevsky2012}
A.~Krizhevsky, I.~Sutskever, and G.E. Hinton,
\newblock ``Advances in neural information processing systems,''
\newblock {\em NIPS}, pp. 1097--1105, 2012.

\bibitem{simonyan2014very}
Karen Simonyan and Andrew Zisserman,
\newblock ``Very deep convolutional networks for large-scale image
  recognition,''
\newblock {\em arXiv preprint arXiv:1409.1556}, 2014.

\bibitem{bebanko2015}
A.~Babenko and V.~Lempitsky,
\newblock ``Aggregating local deep features for image retrieval,''
\newblock {\em Proceedings of the IEEE international conference on computer
  vision}, pp. 1269--1277, 2015.

\bibitem{arandjelovic2016}
R.~Arandjelovic, P.~Gronat, A.~Torii, T.~Pajdla, and J.~Sivic,
\newblock ``{NetVLAD}: {CNN} architecture for weakly supervised place
  recognition,''
\newblock {\em Proceedings of the IEEE Conference on Computer Vision and
  Pattern Recognition}, pp. 5297--5307, 2016.

\bibitem{gordo2016}
A.~Gordo, J.~Almazán, J.~Revaud, and D.~Larlus,
\newblock ``Deep image retrieval: Learning global representations for image
  search.,''
\newblock {\em European conference on computer vision}, pp. 241--257, 2016.

\bibitem{lowe2004}
D.G. Lowe,
\newblock ``Distinctive image features from scale-invariant keypoints,''
\newblock {\em International journal of computer vision}, pp. 91--110, 2004.

\bibitem{perronnin2010}
F.~Perronnin, J.~Sanchez, and T.~Mensink,
\newblock ``Improving the fisher kernel for large-scale image classification,''
\newblock {\em European conference on computer vision}, pp. 143--156, 2010.

\bibitem{jegou2011}
H.~Jegou, M.~Douze, and C.~Schmid,
\newblock ``Product quantization for nearest neighbor search,''
\newblock {\em IEEE transactions on pattern analysis and machine intelligence},
  pp. 117--128, 2011.

\bibitem{duan2014}
L.Y. Duan, J.~Lin, J.~Chen, T.~Huang, and W.~Gao,
\newblock ``Compact descriptors for visual search,''
\newblock {\em IEEE Multimedia}, pp. 30--40, 2014.

\bibitem{ge2014}
T.~Ge, K.~He, Q.~Ke, and J.~Sun,
\newblock ``Optimized product quantization,''
\newblock {\em IEEE transactions on pattern analysis and machine intelligence},
  pp. 744--755, 2014.

\bibitem{martinez2018}
Julieta Martinez, Shobhit Zakhmi, Holger~H Hoos, and James~J Little,
\newblock ``Lsq++: Lower running time and higher recall in multi-codebook
  quantization,''
\newblock in {\em Proceedings of the European Conference on Computer Vision
  (ECCV)}, 2018, pp. 491--506.

\bibitem{cevikalp2017}
H.~Cevikalp, M.~Elmas, and S.~Ozkan,
\newblock ``Large-scale image retrieval using transductive support vector
  machines,''
\newblock {\em Computer Vision and Image Understanding}, 2017.

\bibitem{morozov2019}
Stanislav Morozov and Artem Babenko,
\newblock ``Unsupervised neural quantization for compressed-domain similarity
  search,''
\newblock in {\em Proceedings of the IEEE International Conference on Computer
  Vision}, 2019, pp. 3036--3045.

\bibitem{araujo2018}
Hyeonwoo Noh, Andre Araujo, Jack Sim, Tobias Weyand, and Bohyung Han,
\newblock ``Large-scale image retrieval with attentive deep local features,''
\newblock {\em Proceedings of the IEEE international conference on computer
  vision}, 2017.

\bibitem{araujo2014}
A.~Araujo, J.~Chaves, D.~Chen, R.~Angst, and B.~Girod,
\newblock ``Stanford {I2V}: a news video dataset for query-by-image
  experiments,''
\newblock {\em Proceedings of the 6th ACM Multimedia Systems Conference}, pp.
  237--242, 2014.

\bibitem{sivic20132}
A.~Torii, J.~Sivic, T.~Pajdla, and M.~Okutomi,
\newblock ``Visual place recognition with repetitive structures,''
\newblock {\em Proceedings of the IEEE Conference on Computer Vision and
  Pattern Recognition}, pp. 883--890, 2013.

\bibitem{araujo2017}
A.~Araujo and B.~Girod,
\newblock ``Large-scale video retrieval using image queries,''
\newblock {\em IEEE Transactions on Circuits and Systems for Video Technology},
  2017.

\bibitem{aran2012}
R.~Arandjelovic and A.~Zisserman.,
\newblock ``Three things everyone should know to improve object retrieval,''
\newblock {\em Proceedings of the IEEE Conference on Computer Vision and
  Pattern Recognition}, pp. 2911--2918, 2012.

\bibitem{chum2011}
O.~Chum, A.~Mikulik, M.~Perdoch, and J.~Matas,
\newblock ``Total recall {II}: Query expansion revisited,''
\newblock {\em Proceedings of the IEEE Conference on Computer Vision and
  Pattern Recognition}, pp. 889--896, 2011.

\bibitem{de2016large}
Gabriel de~Oliveira~Barra, Mathias Lux, and Xavier Giro-i Nieto,
\newblock ``Large scale content-based video retrieval with livre,''
\newblock in {\em 2016 14th International Workshop on Content-Based Multimedia
  Indexing (CBMI)}. IEEE, 2016, pp. 1--4.

\bibitem{garcia2018asymmetric}
Noa Garcia and George Vogiatzis,
\newblock ``Asymmetric spatio-temporal embeddings for large-scale
  image-to-video retrieval,''
\newblock BMVC, 2018.

\end{thebibliography}

\end{document}